\documentclass[]{fairmeta}

\usepackage[table]{xcolor}
\usepackage{xcolor}
\usepackage{xspace}
\usepackage{enumitem}
\definecolor{mygreen}{HTML}{2E6037}
\definecolor{mygreendiag}{HTML}{00BF63}
\usepackage{pifont}
\usepackage[most]{tcolorbox}
\newcommand{\ours}{\textsc{Ours}\xspace}
\newcommand{\cmark}{\textcolor{green!60!black}{\ding{51}}}  
\newcommand{\xmark}{\textcolor{red}{\ding{55}}}  
\newcommand{\methodname}{EgoAVU}

\newcommand{\traindata}{EgoAVU-\textit{Instruct}}
\newcommand{\evaldata}{EgoAVU-\textit{Bench}}

\title{\methodname: Egocentric Audio-Visual Understanding}

\author[1,2,*]{Ashish Seth}
\author[1]{Xinhao Mei}
\author[1]{Changsheng Zhao}
\author[1]{Varun Nagaraja}
\author[1]{Ernie Chang}
\author[1]{Gregory P. Meyer}
\author[1]{Gael Le Lan}
\author[1]{Yunyang Xiong}
\author[1]{Vikas Chandra}
\author[1]{Yangyang Shi}
\author[2]{Dinesh Manocha}
\author[1,\dagger]{Zhipeng Cai}

\affiliation[1]{Meta}
\affiliation[2]{University of Maryland, College Park}

\contribution[*]{Work done during intership at Meta}
\contribution[\dagger]{Project Lead}

\abstract{Understanding egocentric videos plays a vital role for embodied intelligence. Recent multi-modal large language models (MLLMs) can accept both visual and audio inputs. However, due to the challenge of obtaining text labels with coherent joint-modality information, whether MLLMs can jointly understand both modalities in egocentric videos remains under-explored. To address this problem, we introduce \methodname, a scalable data engine to automatically generate egocentric audio-visual narrations, questions and answers. \methodname\ enriches human narrations with multimodal context and generates audio-visual narrations through cross-modal correlation modeling. Token-based video filtering and modular, graph based curation ensure both the data diversity and quality. Leveraging \methodname, we construct \traindata\ --- a large scale training dataset of 3M samples, and \evaldata\ --- a manually verified evaluation split covering diverse tasks. 
\evaldata\ clearly reveals the limitation of existing MLLMs: they bias heavily towards visual signals, often neglecting audio cues or failing to correspond audio with the visual source. Finetuning MLLMs on \traindata\ effectively solves this issue, enabling up to 113\% performance improvement on \evaldata. Such benefit can also transfer to other benchmarks such as EgoTempo and EgoIllusion, achieving up to 28\% relative performance gain. Code will be released to the community.}

\date{\today}
\correspondence{Ashish Seth: \email{aseth125@umd.edu}, Zhipeng Cai: \email{czptc2h@gmail.com}}

\metadata[Project Page]{\url{https://cs20s030.github.io/EgoAVU/}}
\metadata[Code]{\url{https://github.com/facebookresearch/EgoAVU}}
\metadata[Data]{\url{https://huggingface.co/datasets/facebook/EgoAVU_data}}

\begin{document}

\maketitle

\section{Introduction}
\label{sec:intro}

Egocentric videos capture rich and dynamic first person audio–visual information centered on daily human activities, such as cooking, painting, or assembling objects~\citep{grauman2022ego4d,nasirimajd2023epic,grauman2024ego}. Understanding such data plays a vital role in embodied intelligence and mixed-realities~\citep{martins2023extending,nagarajan2023egoenv,puig2020watch, savva2019habitat}. 

The highly dynamic camera motion and the limited field of view make comprehensive egocentric video understanding challenging with solely visual cues~\citep{chen2024action2soundambientawaregenerationaction}. This motivates the use of audio information, which provides persistent contextual signals tied to ongoing events. Though recent multi-modal language models (MLLMs) can accept both audio and visual inputs~\citep{ xu2025qwen3, xu2025qwen2, team2024gemini, han2023imagebindllmmultimodalityinstructiontuning, MiniCPM2024}, \textit{whether they understand the joint dynamic of audio–visual signals in egocentric videos} remains an open question.

The main bottleneck to study this problem arguably is limited data. For training, existing egocentric datasets, such as MultiHop-EgoQA~\citep{chen2025grounded} and MM-Ego~\citep{ye2024mm}, are derived mostly from the human narrations in Ego4D~\citep{grauman2022ego4d}. While these narrations provide valuable human supervision, they bias toward describing human–object interactions and lack broader environmental context or the diversity of auditory signals in egocentric recordings. For evaluation, existing benchmarks~\citep{mangalam2023egoschema, cheng2024egothink, plizzari2025omnia} focus mainly on visual cues, limiting their ability to assess integrated audio–visual reasoning. Though exocentric audio–visual benchmarks~\citep{yang2022avqa, li2024omnibench, sung2024avhbench, ma2024look} exist, their multi-modal dynamics are fundamentally different. 


To address these problems, we introduce \methodname, a fully automated data engine that can generate diverse and high quality audio-visual-language data from public egocentric datasets such as Ego4D~\citep{grauman2022ego4d}. \methodname\ comprises four key components: (i) \textit{Enhancing egocentric narrations} by enriching human descriptions with environmental context, visual object details, and audio captions generated using diverse open-source MLLMs~\citep{bai2025qwen25vltechnicalreport, xu2025qwen2, llama3modelcard}; (ii) \textit{Filtering videos with rich audio–visual dynamics} by selecting egocentric clips featuring varied action sequences, human–object interactions, and a wide range of ambient and foreground sounds; (iii) \textit{Generating fine-grained event captions through audio–visual correlation} by integrating modality-specific cues such as actions, objects, and sounds, and modeling their relationships to enhance multimodal reasoning; and (iv) \textit{Curating diverse audio–visual understanding tasks} that span grounding, temporal reasoning, scene understanding, and audio–visual hallucination.

Leveraging \methodname, we construct \traindata\ --- a large scale training dataset of 9K egocentric videos with 3M audio–visual-language samples, and \evaldata\ --- an evaluation split of 900 videos containing 3K manually verified samples. The generated data features strong multi-modal correspondence, long average video durations (4min), open and close-ended questions capturing diverse aspects of egocentric audio-visual understanding. 

Comprehensive experiments show that existing MLLMs perform poorly on \evaldata, revealing significant limitations in their joint audio–visual reasoning capabilities. The seven models we tested show consistent bias towards the vision modality, often neglecting audio cues or struggle to connect them to the correct visual source. Fine-tuning MLLMs such as Qwen2.5-Omni~\citep{xu2025qwen2} on \traindata\ can effectively close this gap, resulting in up to 113\% relative performance boost on \evaldata. More importantly, the performance gain is transferrable to other egocentric benchmarks, such as EgoTempo~\citep{plizzari2025omnia} and EgoIllusion~\citep{seth-etal-2025-egoillusion}, achieving up to 28\% relative performance improvement.



\begin{figure*}[t]
    \includegraphics[width=1.0\linewidth]{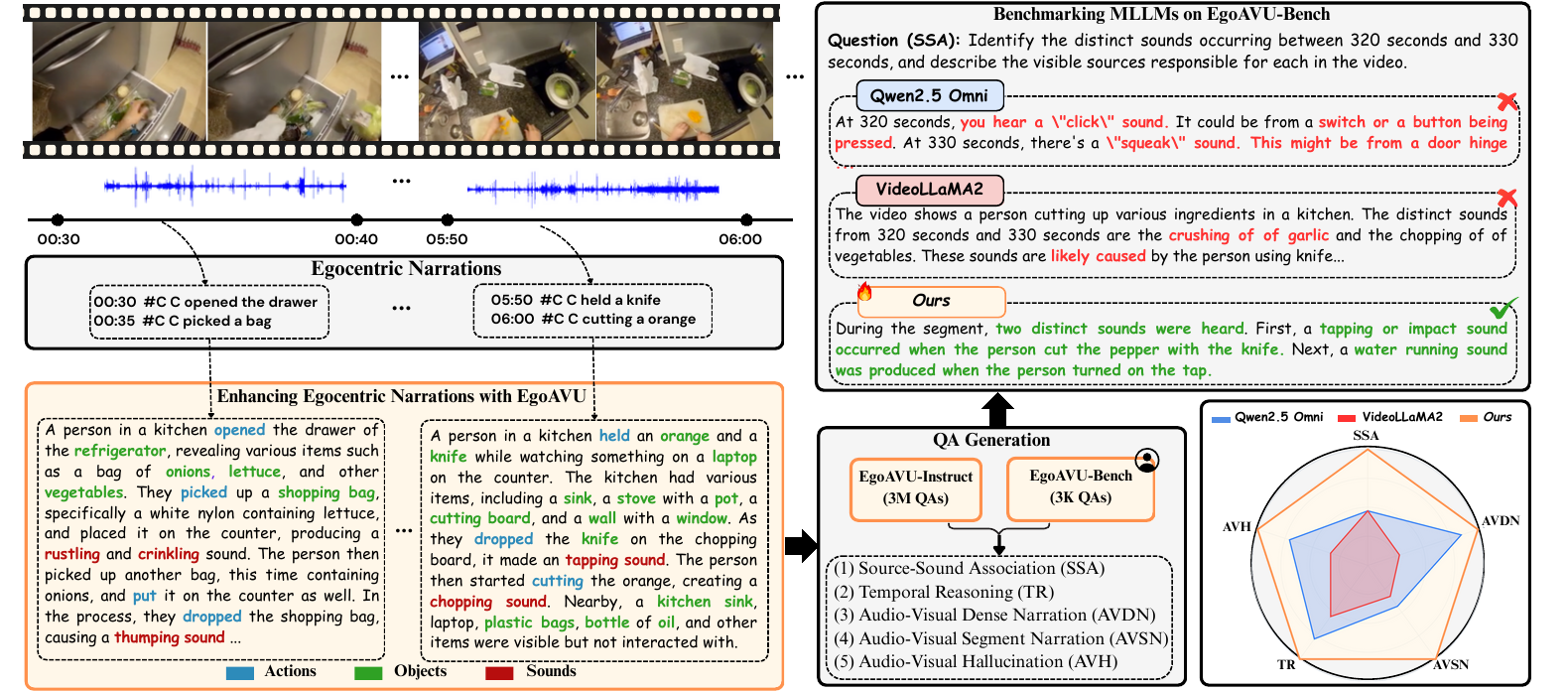}
    \caption{\small \textbf{Overview of \methodname.} We introduce \methodname, a scalable and automated data engine to enable egocentric audio–visual understanding. \methodname \ enriches existing egocentric narrations by integrating human actions with environmental context, explicitly linking visible objects and the sounds produced during interactions or surroundings. Leveraging this pipeline, we construct \emph{\traindata\ (3M QAs)} and \emph{\evaldata\ (3K verified QAs)}, enabling systematic training and evaluation of MLLMs. Models finetuned with \traindata\ exhibit high audio-visual grounding in egocentric settings. \vspace{1em}}
\end{figure*}

\section{Related Work}

\noindent{\textbf{Multimodal-Large Language Models.}} Recent advances in Multimodal Large Language Models (MLLMs)~\citep{team2024gemini, achiam2023gpt, xu2025qwen3, xu2025qwen2, wang2024qwen2, MiniCPM2024, cheng2024videollama} have substantially extended the capabilities of large language models~\citep{achiam2023gpt, llama3modelcard, yang2025qwen3} beyond text, enabling unified understanding of visual and auditory inputs. Despite this progress, existing MLLMs face critical limitations when applied to egocentric audio-visual understanding. First, leading models developed for egocentric video understanding, such as MM-Ego~\citep{ye2024mm} and EgoVLPv2~\citep{pramanick2023egovlpv2}, lack the ability to incorporate audio cues, limiting them to perform visual only tasks. Second, even models capable of handling viusal and auditory signals, including Qwen2.5-Omni~\citep{xu2025qwen2} and Video-LLaVA2~\citep{cheng2024videollama}, remain primarily trained and benchmarked on exocentric audio-visual data~\citep{yang2022avqa, geng2025longvale}. As a result, they struggle to generalize to egocentric settings, which exhibit fundamentally different characteristics such as dynamic camera motion, frequent self-occlusions, and distinct audio profiles. These limitations highlight a \textit{significant gap in current MLLMs’ to jointly understand audio–visual cues in egocentric videos.} 

\noindent{\textbf{Egocentric-Video Understanding.}} The field of egocentric video understanding has gained increasing attention due to its relevance in augmented reality (AR)~\citep{nagarajan2023egoenv, martins2023extending} and embodied AI~\citep{savva2019habitat, puig2020watch}. Large-scale datasets such as Ego4D~\citep{grauman2022ego4d}, EPIC-KITCHENS~\citep{nasirimajd2023epic}, and Ego-Exo4D~\citep{grauman2024ego} have driven progress by enabling the creation of egocentric video–language corpora (e.g., MultiHop-EgoQA~\citep{chen2025grounded}, EgoTextVQA~\citep{zhou2025egotextvqa}, QaEgo4D~\citep{barmann2022did}) and benchmarks (e.g., EgoSchema~\citep{mangalam2023egoschema}, EgoTempo~\citep{plizzari2025omnia}, EgoIllusion~\citep{seth-etal-2025-egoillusion}). However, these datasets are predominantly constructed from textual narrations, focusing on human–object interactions while overlooking environmental and auditory cues, critical for accurately inferring actions, contextual dynamics, and scene semantics in egocentric settings. Recent benchmarks such as EgoTempo and EgoIllusion attempt to enhance contextual understanding through visual captioning, but their reliance on closed-source models (e.g., Gemini~\citep{team2024gemini}, GPT-4o~\citep{achiam2023gpt}) makes large-scale, reproducible data generation challenging. In this work, we address these problems by \textit{introducing a scalable data-engine capable of generating time-aligned joint audio-visual generation while utilizing open-source MLLMs~\citep{xu2025qwen2, bai2025qwen25vltechnicalreport,llama3modelcard}}.

\section{Method}

\begin{figure*}[t]
    \centering
    \includegraphics[width=1.0\linewidth]{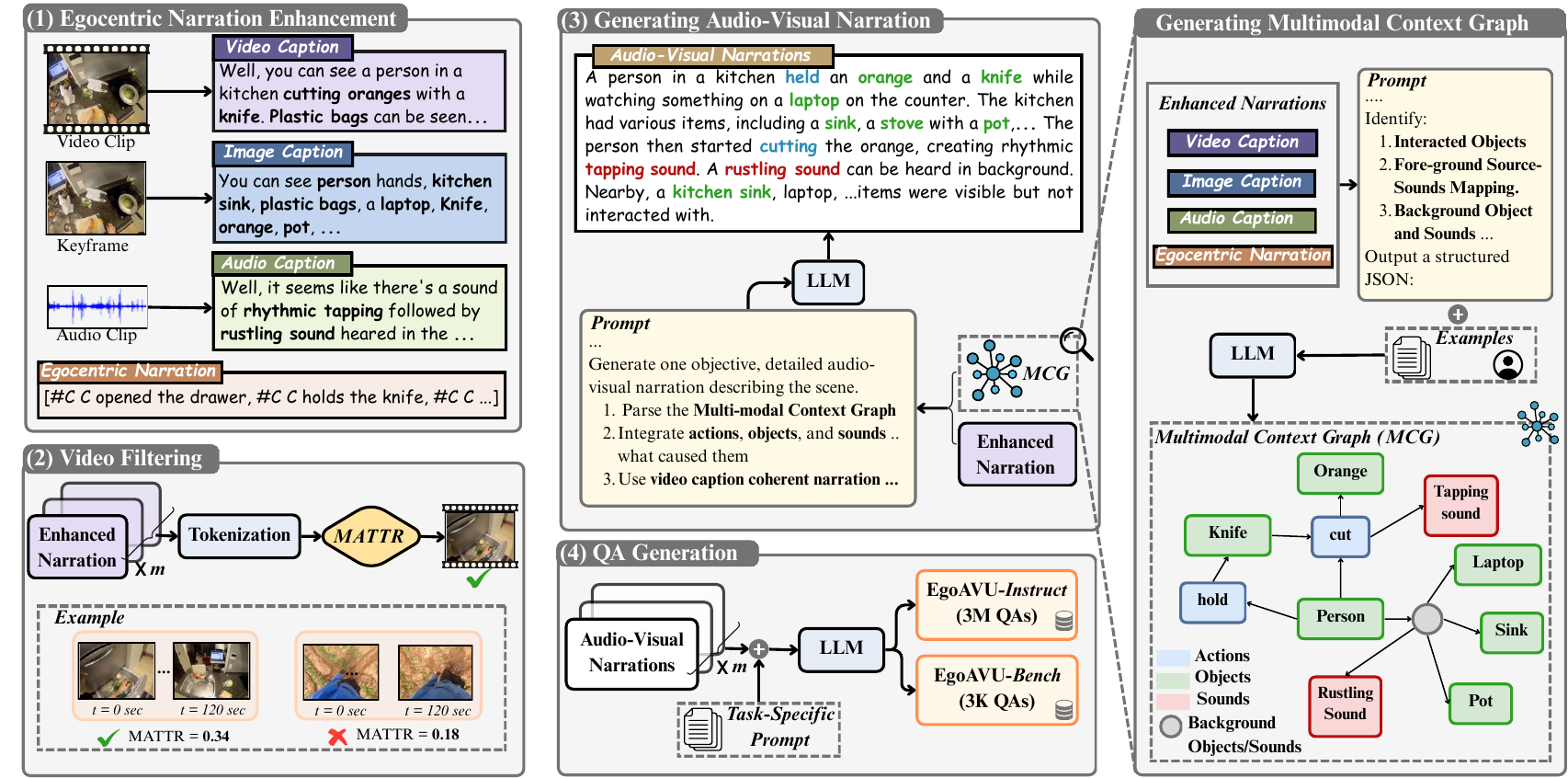}
    \caption{\textbf{\methodname\ pipeline.} \methodname\ consists of four key components. (1) For each egocentric video clip, \methodname\ enhances the raw narration with detailed multisensory context using open-source MLLMs~\citep{bai2025qwen25vltechnicalreport,xu2025qwen2}. (2) These enriched narrations are then used to select clips that exhibit diverse audio–visual dynamics. (3) Next, \methodname\ constructs a Multimodal Context Graph (MCG), automatically generated via open-source LLMs~\citep{llama3modelcard}, to capture complex cross-modal relations. The MCG is parsed alongside the enhanced narrations to produce coherent audio–visual narrations. (4) The generated audio-visual narrations are leveraged to generate high-quality audio–visual QA pairs, forming both the instruction-tuning dataset \traindata\ and the evaluation benchmark \evaldata.}
    \label{fig:method_diag}
\end{figure*}

Fig.\ref{fig:method_diag} provides an overview of \methodname. We begin by collecting and organizing egocentric data (Sec.\ref{sec:data_collection}). Next, we enhance egocentric narrations using MLLM-generated multi-modal context (Sec.\ref{sec:enh_ego_narration}), and use the tokens of enriched narration to filter videos for temporal diversity (Sec.\ref{sec:video_filter}). We then introduce a multi-modal context graph (MCG) that captures complex cross-modal relations, and parse these graphs with open-source LLMs to fuse multi-modal information into a single dense narration (Sec.\ref{sec:narration_gen}). Finally, this fused narration is used to generate QA pairs for joint audio-visual understanding (Sec.\ref{sec:QA_gen}). To ensure the scalability, \methodname\ only utilizes open source models. Please refer to Appendix~\ref{appdx:video_filter} for example outputs and prompts for each module.

\subsection{Data Collection}
\label{sec:data_collection}
We begin by collecting videos from the Ego4D dataset~\citep{grauman2022ego4d}, and filter out videos that lack audio tracks. Each video in Ego4D is accompanied by a set of narrations that provide first-person descriptions of the events, such as ``\#C C holds a cup,'' where ``\#C C'' refers to the camera wearer. These narrations can be represented as $\{N_{j}, t_{j}\}_{j=1}^{K}$, where $N_{j}$ denotes the narration text, $t_{j}$ is the corresponding timestamp, and $K$ is the total number of narrations in the video. To obtain the temporal boundries of each narration, we use the strategy proposed in prior works~\citep{plizzari2025omnia, lin2022egocentric}, where temporal boundries $T_{j}$ for narration $N_{j}$ is defined as:
\begin{equation}
    T_j = \left\{t_j \ - \ \frac{\beta_i}{2\alpha},  t_j \ + \ \frac{\beta_i}{2\alpha}\right\}.
\end{equation}
$\beta_i$ represents the average interval between consecutive timestamps ($\{T_j, T_{j+1}\}$), and $\alpha$ denotes the global average of $\beta_i$ across the entire dataset. Since the segments associated with each narration are short (on average 3s), they are insufficient to capture fine-grained visual and auditory information. Following~\citep{di2024grounded}, for each video $v$, we group consecutive segments and their corresponding narrations to form video clips $v_j$ of at least 10s and not more than 360s. 

\subsection{Narration Enhancement}
\label{sec:enh_ego_narration}

After data collection, we generate descriptions about audio and video frames to enrich the original egocentric action-centric narrations. The goal of this stage is to obtain detailed descriptions for both modalities, which paves the way for diverse and fine-grained QA generation. The most straight-forward approach is to input both video frames and audio tracks into an MLLM, and prompt it to generate the audio and visual descriptions. However, our initial experiments show that MLLMs, when provided with audio and video inputs together, cannot capture important details due to modality bias and hallucination.

Specifically, we evaluate open-source models, including Qwen2.5-Omni~\citep{xu2025qwen2} and MiniCPM-o~\citep{MiniCPM2024}, on 200 randomly sampled video clips. Each model is first used in a uni-modal setting, captioning visual and auditory information independently, i.e., visual captioning without audio input and audio captioning without video input. We then perform joint audio–visual captioning by providing both modalities simultaneously. After that, we manually compare the consistency for both visual and audio modalities by computing the ratio where an object/event captured in the uni-modal output is also correctly appearing in the multi-modal output. I.e., whether the multi-modal output can capture all the details as in the uni-modal settting.

We observe a consistent pattern where models either omit various sounds or associate audio cues with incorrect visual events. For example, Qwen2.5-Omni shows an error rate of $54.3\%$ for audio and $25.4\%$ for visual consistency, while MiniCPM-o yields $68.2\%$ and $31.2\%$, respectively (see Fig~\ref{fig:mcg_qualitative} for example). These findings show that existing MLLMs struggle to maintain accurate and detailed cross-modal grounding in egocentric contexts, motivating a modular data engine that leverages specialized models to process different modalities independently.


With this observation, we leverage a collection of MLLMs to extract rich multisensory information from individual modalities separately. Specifically, for each video segment, we first capture detailed spatial descriptions of objects by applying an image captioner such as Qwen2.5-VL~\citep{bai2025qwen25vltechnicalreport} to the center frame. To model temporal dynamics, including camera motion, action sequences, and auditory events, we employ Qwen2.5-Omni in two complementary modes. First, we use it as a video captioner, processing video frames without audio, to generate coherent video-level narrations. Then, we use the same model as an audio captioner without visual inputs to produce detailed auditory narrations that capture both foreground sounds closely tied to human activities (e.g., impacts and hissing) and background sounds (e.g., bird chirping, wind blowing). This process results in time-aligned uni-modal narrations that encapsulate both visual and auditory aspects of the scene.
   
\subsection{Video Filtering for Diversity Enhancement}\label{sec:video_filter}
After obtaining the enhanced narrations, we use them to compute lexical diversity, filtering videos with rich and diverse visual and auditory signals. Specifically, for each video $v$, we combine the segment-level narrations into a single narration that captures all objects, actions, and sounds present in the video. We then tokenize this narration into $\mathcal{T}_v = \{t_1, \dots, t_n\}$ and compute the Moving-Average Type–Token Ratio (MATTR)~\citep{covington2010cutting}, which averages the proportion of unique words across sliding windows of size $w$:
\begin{equation}
\text{MATTR}(\mathcal{T}_v) = \frac{1}{n - w + 1}
\sum_{i=1}^{n-w+1}
\frac{ \left| \text{Uni.}(t_i, \dots, t_{i+w-1}) \right| }{w}.
\end{equation}
Higher MATTR scores indicate narrations that describe a wider variety of objects, actions, and auditory events. We retain videos whose MATTR exceeds a threshold $\tau=0.3$, effectively removing the bottom 25\% of our distribution to filter static or repetitive descriptions, leading to the final video count of 9,900. 

\subsection{Audio-Visual Narration Generation} \label{sec:narration_gen}
Our next objective is to generate a unified narration that captures both auditory and visual information for each segmented video clip. We do this leveraging text-based LLMs. When synthetically merging unimodal narrations, the LLM must first implicitly retrieve key details across modalities, such as which objects the human interacts with, which remain in the background, and what actions or objects produce specific sounds. It must then integrate this information into a coherent and perceptually grounded narration. However, our preliminary experiments show that directly prompting open-source LLMs such as LLaMA-70B~\citep{llama3modelcard} often fails to maintain consistent human–object interactions and sound–source associations in their responses (See Appendix~\ref{appdx:MCG_ablation} for detailed comparison).

We design a two-stage pipeline that addresses this challenge. First, we organize audio-visual cues from uni-modal narrations into a structured representation called the \emph{Multi-modal Context Graph (MCG)}. As shown in Fig.~\ref{fig:method_diag}, MCG captures relationships between visible actions, objects, and audible sounds.
Each MCG is generated by prompting an LLM (LLaMA-70B) with enhanced narrations to extract the following information :
\begin{itemize}
\item \textit{Interacted Objects:} Objects with which the person physically interacts, along with interaction types, inferred from action narrations.
\item \textit{Background Objects:} Objects visible in the environment but never interacted with, identified by comparing objects and action narrations.
\item \textit{Foreground Sounds:} Human-induced action sounds correlated with specific actions (e.g., \textit{impact sound} $\rightarrow$ ``C places the phone on the table''), or ambient sounds grounded in visible scene elements (e.g., ``dog barking'' aligned with a visible dog).
\item \textit{Background Sounds:} Sounds present in the audio track whose sources cannot be visually grounded.
\end{itemize}
We find that MCG makes multi-modal relationships explicit and easily inferable, which we utilize further to generate joint audio-visual narrations.  


Next, we leverage MCGs to guide the generation of unified, high-coherence audio–visual narrations. The goal is to fuse both visual and auditory details into the same multi-modal coherent narration. As shown in Fig~\ref{fig:method_diag}, we provide the LLM (LLaMA-70B) with enhanced narrations and MCGs, and prompt it to generate the combined narration. The prompt asks the LLM to first extract explicit cues from the MCG, including interacted and background objects, grounded sound events, and their associations with actions and visible sources, and then align these cues with the corresponding temporal descriptions from the video and action-level narrations.  


\begin{table*}[t]
\centering
\footnotesize
\setlength{\tabcolsep}{3pt}
\begin{tabular}{|p{6cm}|p{10cm}|}
\toprule
\textbf{Category} & \textbf{Example} \\
\midrule
\rowcolor{gray!10}
\multicolumn{2}{|c|}{\textit{Open-Ended}} \\[2pt]
\midrule
\textit{Source–Sound Association (SSA)} &
 What is the source of the sizzling sound heard in the video?\\
\midrule
\textit{Audio–Visual Segment Narration (AVSN)} &
 Between 240 and 250 seconds, describe the person's surroundings, actions, and the sounds that can be heard?\\
\midrule
\textit{Audio–Visual Dense Narration (AVDN)} &
 Describe in detail what the person sees, hears, and does throughout the video.\\
\midrule
\rowcolor{gray!10}
\multicolumn{2}{|c|}{\textit{Close-Ended}} \\[2pt]
\midrule
\textit{Temporal Reasoning (TR)} &
 What happened before the person opened a kitchen cabinet?. Choose the correct option from the following options: ... \\
\midrule
\textit{Audio–Visual Hallucination (AVH)} &
 Is there a beeping sound coming from the microwave in the video?\\
\bottomrule
\end{tabular}
\captionof{table}{\textbf{Question Prompts.} Examples of open-ended and close-ended questions in \traindata\ and \evaldata.}
\label{tab:qa_examples}
\end{table*}

\begin{table*}[t]
\centering
\begin{minipage}[t]{0.46\textwidth}
\centering
\includegraphics[width=1\linewidth]{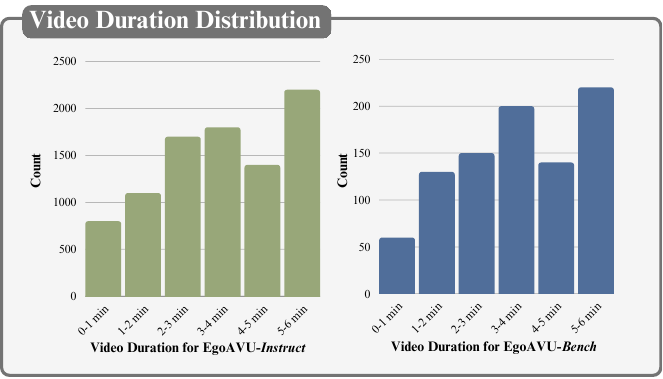}
\captionof{figure}{\textbf{Video duration distribution.} Our videos includes both short clips within 1 min and long videos of 6 min.}
\label{fig:data_stats}

\end{minipage}
\hfill
\begin{minipage}[t]{0.52\textwidth}
\vspace{-10em}

\small
\setlength{\tabcolsep}{4pt}
\resizebox{\columnwidth}{!}{
\begin{tabular}{lcccccc}
\toprule
\textbf{Dataset} & \textbf{A\&V} & \textbf{\# Test} & \textbf{Avg. Dur.} & \textbf{QA-Type} & \textbf{\# Ans.} \\
\midrule
EgoTaskQA~\citep{jia2022egotaskqa} & \xmark & 8k & 25s & Open & 13 \\
EgoSchema~\citep{mangalam2023egoschema} & \xmark & 500 & 3 min & Close & - \\
EgoThink~\citep{cheng2024egothink} & \xmark & 750 & 45s & Open & 4 \\
EgoTempo~\citep{plizzari2025omnia} & \xmark & 500 & 45s & Open & 10 \\
EgoIllusion~\citep{seth-etal-2025-egoillusion} & \xmark & 8k & 3 min & Open+Close & 3 \\
\midrule
\rowcolor{gray!10}
\textbf{\evaldata (Ours)} & \cmark & 3k & 4 min & Open+Close & 200 \\
\bottomrule
\end{tabular}
}
\captionof{table}{\textbf{Benchmark statistics.} Beside having QAs with high audio-visual coherence, \evaldata\ features large number of QA pairs, longer egocentric videos with audio track, support both open/closed QA type and consists of significantly longer and descriptive responses.}
\label{tab:benchmark_comparison}
\end{minipage}
\end{table*}

\begin{figure*}[t]
\centering
\begin{minipage}[t]{0.48\textwidth}
\centering
\includegraphics[width=\columnwidth]{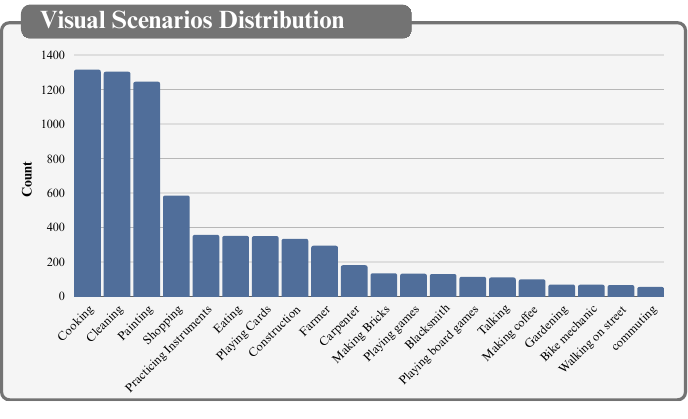}
\caption{\textbf{Distribution of 20 most common visual scenarios in \traindata\ and \evaldata.}}
\label{fig:visual_scenarios}
\end{minipage}
\hfill
\begin{minipage}[t]{0.48\textwidth}
\centering
\includegraphics[width=\columnwidth]{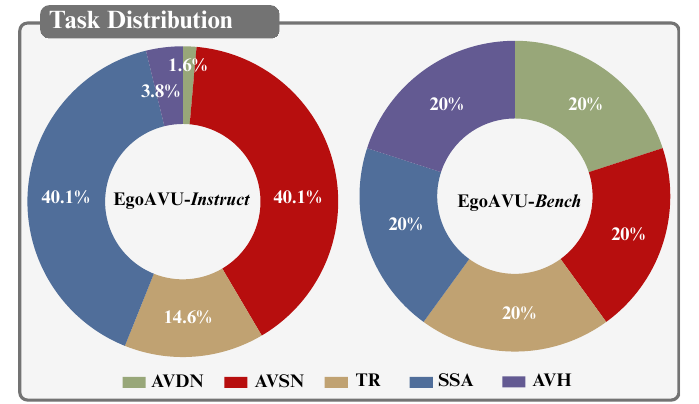}
\caption{\textbf{Distribution of proposed tasks across \traindata\ and \evaldata}.}
\label{fig:task_distribution}
\end{minipage}
\end{figure*}

\subsection{QA Generation}\label{sec:QA_gen}
Using the unified audio–visual narrations, \methodname\ synthesizes question–answer pairs that probe diverse aspects of egocentric audio-visual understanding.

\noindent{\textbf{\methodname\ Taxonomy.}} 
We design five categories of QAs encompassing both \emph{open-ended} questions, where models must produce descriptive responses integrating visual and auditory cues, and \emph{closed-ended} questions, which test precise multi-modal perceptual understanding through multiple-choice or binary (Yes/No) formats. A detailed description of each category is provided below.



\begin{itemize}
    \item \textbf{Open-ended QAs.} These questions are primarily free-form, requiring the model to develop a comprehensive understanding of temporal and spatial dynamics across visual and auditory cues in egocentric videos. The specific categories include: (1) \textit{Sound–Source Association (SSN):} Identify various foreground sounds in the video and determine their corresponding visible sources including human actions or various objects shown in the video. (2) \textit{Audio–Visual Segment Narration (AVSN):} Answer segment-level questions by producing coherent and natural narrations that describe what the person is doing, seeing, and hearing, including both foreground and background sounds, grounded within the specified temporal range. (3) \textit{Audio–Visual Dense Narration (AVDN):} Extend previous task to the entire video, assessing the model’s ability to maintain narrative coherence across the complete video.  
    \item \textbf{Close-ended QAs.} These questions are particularly useful for \textit{adversarial testing}. Close-ended QAs allow the construction of fine-grained distractors and counterfactuals to probe a model’s susceptibility to spurious correlations in audio-visual understanding. We design two main categories: (1) \textit{Temporal Reasoning (TR):} Multiple-choice questions with four options that assess the model’s understanding of temporal relationships among multi-modal events such as human actions, visual objects, and auditory cues in egocentric videos. These questions focus on reasoning about event ordering, for example identifying which event occurred first or last, or answering before/after queries. (2) \textit{Audio–Visual Hallucination (AVH):} Binary (``Yes''/``No'') questions that evaluate the model’s tendency to hallucinate when verifying the presence of actions, objects, or sounds throughout the video.

\end{itemize}

\subsection{Dataset Overview}\label{sec:statistics}
Leveraging \methodname, we construct the first egocentric audio–visual dataset suite for training and evaluating MLLMs. Our large-scale instruction-tuning corpus, \traindata, comprises approximately 3M samples with 9K egocentric videos, while the evaluation benchmark, \evaldata, contains 3K QA pairs with 900 distinct videos. As shown in Fig.~\ref{fig:visual_scenarios}, both \traindata\ and \evaldata\ encompass a wide range of real-world visual scenarios, such as cooking, painting, and other indoor/outdoor activities. Extensive manual verifications are conducted on \evaldata\ to ensure that the audio-visual information is correctly grounded. As illustrated in Fig.~\ref{fig:task_distribution}, both \traindata and \evaldata\ covers all five question categories introduced earlier, featuring a balanced mix of open- and closed-ended formats. As shown in Fig.~\ref{fig:data_stats}, the video length in both datasets vary from 1 to 6 minutes, providing rich temporal diversity for model training and evaluation. 

Table~\ref{tab:benchmark_comparison} compares existing egocentric benchmarks with our \evaldata. 
Besides filling the gap of egocentric audio-visual understanding, \evaldata\ includes a large number of QAs, features longer videos with synchronized audio tracks, supports both open- and closed-ended question formats, and provides significantly longer and more descriptive responses.

\label{sec:benchmark}

\begin{table*}[t]
\centering
\resizebox{1.0\textwidth}{!}{
\begin{tabular}{lccccccccccc}
\toprule
\textbf{Models} & \textbf{Size} & \textbf{SSA} & 
\multicolumn{3}{c}{\textbf{AVDN}} & 
\multicolumn{3}{c}{\textbf{AVSN}} & 
\textbf{TR} & \textbf{AVH} \\
\cmidrule(lr){4-6} \cmidrule(lr){7-9}
 &  & \textbf{S} ($\uparrow$) & \textbf{S} ($\uparrow$) & \textbf{M} ($\uparrow$) & \textbf{R} ($\uparrow$) 
 & \textbf{S} ($\uparrow$) & \textbf{M} ($\uparrow$) & \textbf{R} ($\uparrow$) 
 & \textbf{Acc.} ($\uparrow$) & \textbf{Acc.} ($\uparrow$) \\
\midrule
\rowcolor{gray!10}
\multicolumn{11}{c}{\textit{Open-Source MLLMs}} \\
\midrule
VideoLLaMA2~\citep{cheng2024videollama} & 7B & 1.51 & 1.88 & 3.65 & 8.32 & 1.71 & 7.50 & 13.89 & 37.00 & 20.32 \\
Baichuan-Omni~\citep{li2024baichuan} & 7B & 1.49 & 1.92 & 4.82 & 9.75 & 1.79 & 8.34 & 14.21 & 39.85 & 21.10 \\
Intern-Omni~\citep{chen2024internvl} & 8.7B & 1.47 & 1.95 & 5.20 & 10.11 & 1.82 & 8.69 & 14.70 & 41.22 & 21.75 \\
Phi4-mm~\citep{abouelenin2025phi} & 8B & 1.42 & 1.59 & 8.79 & 13.13 & 1.69 & 12.17 & 16.90 & 45.04 & 22.89 \\
MiniCPM-o~\citep{MiniCPM2024} & 8B & 1.43 & 2.27 & 10.84 & 14.77 & 2.06 & 9.68 & 12.19 & 26.44 & 21.76 \\
Qwen2.5-Omni~\citep{xu2025qwen2} & 3B & 1.45 & 2.17 & 8.55 & 13.45 & 1.85 & 8.63 & 13.08 & 46.40 & 26.28 \\
Qwen2.5-Omni~\citep{xu2025qwen2} & 7B & 1.50 & 2.37 & 10.69 & 14.74 & 1.99 & 9.99 & 13.39 & 53.20 & 42.69 \\
\midrule
\rowcolor{gray!10}
\multicolumn{11}{c}{\textit{MLLM trained with \traindata}} \\
\midrule
\ours$_{({LoRA})}$ & 7B 
& 3.15 
& 2.60
& 12.20
& 17.19
& 2.45
& 22.53
& 28.34
& 64.31
& \textbf{61.69} \\
\ours$_{({Full})}$ & 7B 
& \textbf{3.20} 
& \textbf{2.66} 
& \textbf{12.50} 
& \textbf{17.32} 
& \textbf{2.63} 
& \textbf{22.68}
& \textbf{28.70}
& \textbf{67.84}
& 60.12 \\
\midrule
\rowcolor{gray!10}
\textbf{$\Delta (\%)$} & -- 
& \textcolor{mygreen}{+113.3} 
& \textcolor{mygreen}{+12.2} 
& \textcolor{mygreen}{+16.9} 
& \textcolor{mygreen}{+17.2} 
& \textcolor{mygreen}{+27.6} 
& \textcolor{mygreen}{+86.5} 
& \textcolor{mygreen}{+69.8} 
& \textcolor{mygreen}{+27.2} 
& \textcolor{mygreen}{+30.8} \\
\bottomrule
\end{tabular}
}
\caption{\small \textbf{Main result on \evaldata.} We compare seven MLLMs with joint audio–visual understanding capabilities against our fine-tuned models across a diverse set of tasks in \evaldata, including open-ended QAs: Source-Sound Association (SSA), Audio-Visual Dense Narration (AVDN), and Audio-Visual Segment Narration (AVSN), as well as closed-ended QAs: Temporal Reasoning (TR) and Audio-Visual Hallucination (AVH). For the open-ended tasks, we report LLM-as-Judge (S), METEOR (M), and ROUGE-L (R). For the closed-ended tasks, we report Accuracy (Acc.). Additionally for each task, we compute the relative performance gain ($\Delta$) between the best open-source model and our fine-tuned models.}
\label{tab:av_comparison}
\end{table*}

\section{Results}
\noindent\textbf{Implementation Details.} 
We verify the effectiveness of our instruction-tuning dataset, \traindata, by fine-tuning MLLMs such as Qwen2.5-Omni (7B) on it and compare the finetuned model with baselines on \evaldata. Fine-tuning is performed using LLaMA-Factory~\citep{zheng2024llamafactory}, under two settings: LoRA~\citep{hu2022lora} and full fine-tuning. All experiments are conducted on 64 H100 GPUs with a global batch size of 64, training each model for 5 epochs. To ensure consistent visual coverage across samples, during training, we uniformly sample 300 frames per video. We also uniformly sample data from each of the 5 tasks to achieve a balanced performance (See Appendix~\ref{appdx:implementation} for additional details).



\noindent\textbf{Evaluation Protocol.} For close-ended QAs in \evaldata, we follow \citep{yue2023mmmu} and use regex-based string matching, where we construct robust regular expressions and design a response-processing module to extract key phrases such as option IDs (A, B, C, D), binary indicators (yes/no), and conclusion phrases from long responses for accurate answer matching. For open-ended QAs, similar to prior work~\citep{plizzari2025omnia}, we adopt the LLM-as-a-judge approach, employing Qwen3-235B-A22B-Instruct-2507~\citep{yang2025qwen3} as an open-source judge for reproducibility. The model rates MLLM-generated responses on a 1–5 scale (see Appendix~\ref{appdx:implementation} for the evaluation prompt). We additionally report standard metrics used for dense response evaluation, including ROUGE-L~\citep{lin-2004-rouge} and METEOR~\citep{banerjee2005meteor}.

\begin{figure*}[t]
\centering
\begin{minipage}[t]{0.54\textwidth}
\centering
\vspace{-5.5em} 
\includegraphics[width=1.\columnwidth]{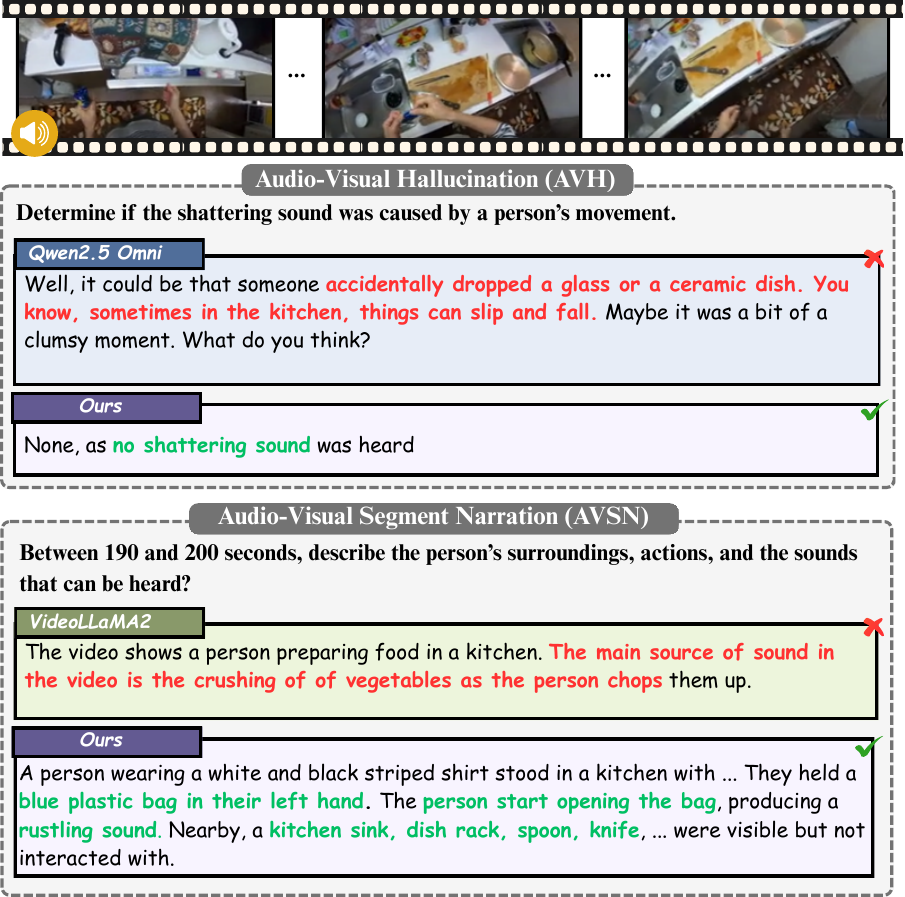}
\captionof{figure}{\centering\textbf{Qualitative Analysis on \evaldata.}}
\label{fig:exp_vis}
\end{minipage}
\hfill
\begin{minipage}[t]{0.45\textwidth}
\centering

\centering
\resizebox{1.0\columnwidth}{!}{%
\begin{tabular}{lcccc}
\toprule
\textbf{Models} & \textbf{Size} & \textbf{Action} & \textbf{Object} & \textbf{Sound} \\
& & \textbf{Acc.} ($\uparrow$) & \textbf{Acc.} ($\uparrow$) & \textbf{Acc.} ($\uparrow$) \\
\midrule
VideoLLaMA2 & 7B & 21.36 & 20.32 & 19.27 \\
Phi4-MM & 8B & 22.45 & 21.74 & 24.49 \\
MiniCPM-o & 8B & 24.49 & 24.46 & 16.33 \\
Qwen2.5-Omni & 3B & 25.00 & 29.35 & 24.49 \\
Qwen2.5-Omni & 7B & 44.39 & 50.00 & 33.67 \\
\midrule
\ours$_{(LoRA)}$ & 7B & 60.20 & 61.09 & 63.78 \\
\ours$_{(Full)}$ & 7B & 61.32 & 62.40 & 64.20\\
\bottomrule
\end{tabular}
}
\captionof{table}{\textbf{Error Analysis for Audio-Visual Hallucination (AVH).}}
\label{tab:avh_comparison}

\vspace{0.3cm} 


\centering
\resizebox{1.0\columnwidth}{!}{%
\begin{tabular}{lcccc}
\toprule
\textbf{Models} & \textbf{Size} & \textbf{Action} & \textbf{Object} & \textbf{Sound} \\
& & \textbf{Acc.} ($\uparrow$) & \textbf{Acc.} ($\uparrow$) & \textbf{Acc.} ($\uparrow$) \\
\midrule
VideoLLaMA2 & 7B & 45.88 & 38.38 & 38.30 \\
Phi4-MM & 8B & 48.81 & 48.48 & 46.74 \\
MiniCPM-o & 8B & 40.48 & 25.00 & 37.50 \\
Qwen2.5-Omni & 3B & 44.44 & 69.70 & 31.91 \\
Qwen2.5-Omni & 7B & 43.53 & 64.65 & 36.17 \\
\midrule
\ours${_{(LoRA)}}$ & 7B & 54.31 & 68.90 & 52.45 \\
\ours${_{(Full)}}$ & 7B & 55.29 & 69.80 & 53.17 \\
\bottomrule
\end{tabular}
}
\captionof{table}{\centering\textbf{Error Analysis for Temporal Reasoning (TR).}}
\label{tab:av_mcq_comparison}

\end{minipage}

\end{figure*}

\begin{figure*}[t]
    \centering
    \includegraphics[width=1.0\textwidth]{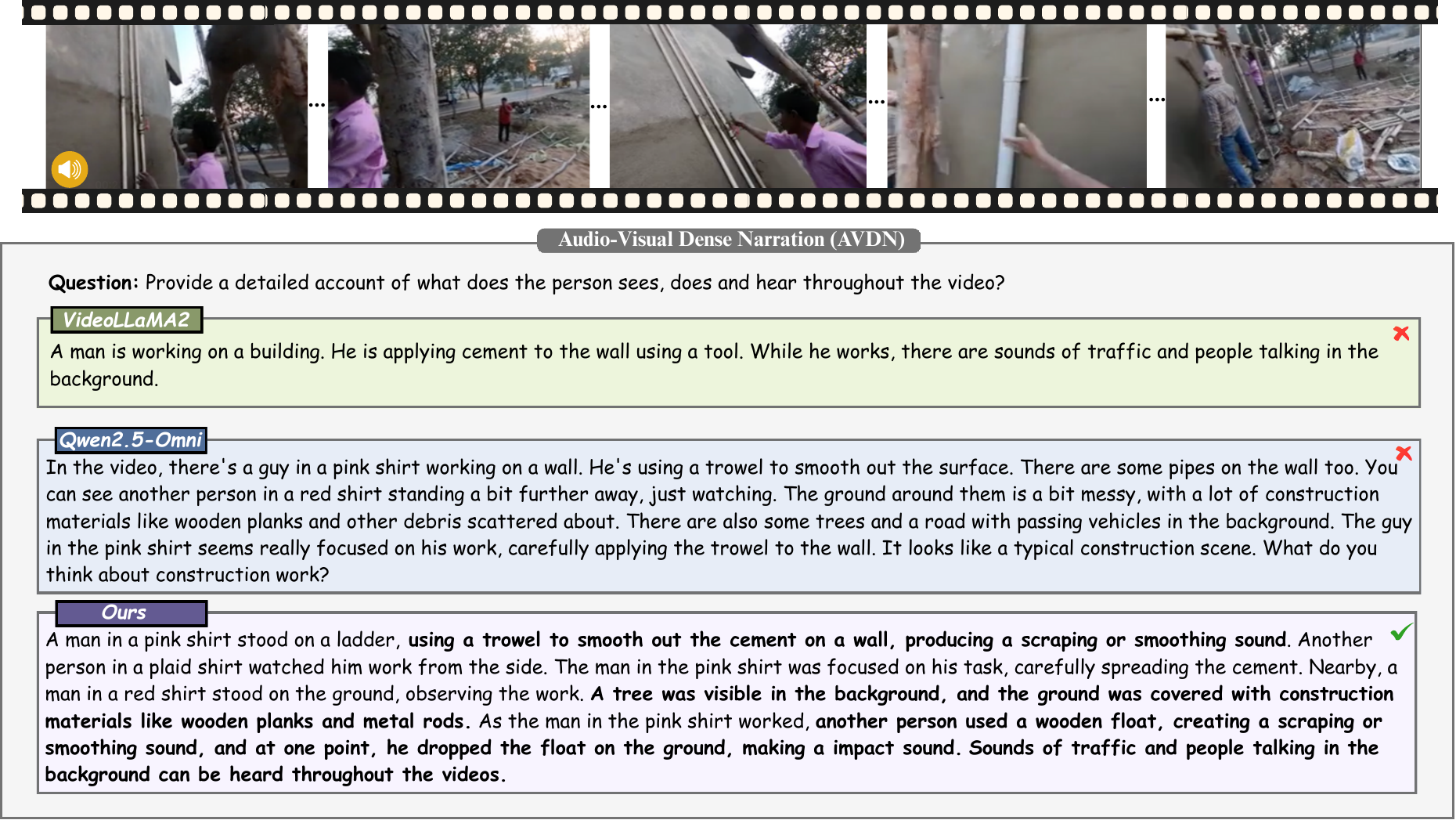}
    \caption{\textbf{Qualitative comparison of various MLLMs on the Audio-Visual Dense Narration (AVDN) task.} Our model fine-tuned on \traindata\ captures significantly more dense visual details than Qwen2.5 Omni and VideoLLaMA2, while also identifying auditory cues related to human actions and background sounds in the video.}
    \label{fig:avdn_qualitative}
\end{figure*}

\subsection{Main Results}

Table~\ref{tab:av_comparison} presents the main result on \evaldata. The key findings are summarized below.
\begin{itemize}
    \item \emph{MLLMs struggle to associate sounds with their visual sources.}  This is evident from their low scores on the Source-Sound Localization (SSA) task, where LLM-as-a-judge (S) evaluations remain below 1.6 out of 5 across all baseline MLLMs. 
    \item \emph{MLLMs struggle to produce coherent and temporally aligned audio-visual narrations.} In both the Audio-Visual Dense Narration (AVDN) and Audio-Visual Segment Narration (AVSN) tasks, even the best-performing model, Qwen2.5-Omni (7B), attains S scores of below 2.4, which is consistent with low caption quality metrics such as ROUGE-L (R) and METEOR (M).
    \item \emph{MLLMs demonstrate limited temporal reasoning over joint audio-visual inputs.} The highest Temporal Reasoning (TR) accuracy, achieved by Qwen2.5-Omni (7B), is merely 53.2\%, with models such as VideoLLaMA2 and MiniCPM-o performing considerably worse. 
    \item \emph{MLLMs frequently hallucinate during audio-visual reasoning}, as reflected by their below 43\% accuracy on the Audio-Visual Hallucination (AVH) task. 
    \item \emph{Finetuning MLLMs on \traindata\ effectively improves egocentric audio-visual understanding.} Our finetuned model yields substantial and consistent performance gains across all tasks. Compared to the best performing baseline, we achieve up to 113.3\% and 44.5\% relative performance improvement on open and close ended tasks respectively. In addition, LoRA and full finetuning both provides considerable performance gain. This shows the possibility to achieve strong audio-visual understanding under resource-limited scenarios.
\end{itemize}


Figure~\ref{fig:exp_vis} compares the responses of VideoLLaMA2 and Qwen2.5-Omni (7B), with our fine-tuned model on both open-ended and close-ended tasks. In close-ended tasks like Audio-Visual Hallucination, when queried about a sound source absent from the video, Qwen2.5-Omni often fabricates a visually plausible yet non-existent source, whereas our model effectively resists such misleading prompts through improved audio–visual grounding. For open-ended tasks such as Audio-Visual Segment Narration, our model exhibits stronger sound–source coherence and more accurate action-sequence understanding. Fig.~\ref{fig:avdn_qualitative} further shows examples on audio-visual dense narration (AVDN). Unlike our fine-tuned model, existing MLLMs produce sparse descriptions for Audio-Visual Dense Narration task in \evaldata\ and further overlook audio cues or fail to ground sounds to their sources.

\subsection{Evaluating Existing Egocentric Benchmarks}
\begin{table}[t]
\centering
\resizebox{\columnwidth}{!}{%
\begin{tabular}{llllll}
\toprule
\textbf{Model} & \textbf{EgoTempo} \textbf{Acc.} ($\uparrow$) & \textbf{EgoIllusion} \textbf{Acc.} ($\uparrow$) & \textbf{EgoSchema} \textbf{Acc.} ($\uparrow$) & \textbf{VideoMME} \textbf{Acc.} ($\uparrow$) & \textbf{AVQA} \textbf{Acc.} ($\uparrow$)\\
\midrule
Qwen2.5-Omni & 16.25 & 56.32 & \textbf{67.43} & \textbf{73.0} & 89.4 \\
\ours$_{(LoRA)}$ & \textbf{20.83}$_{\textcolor{mygreen}{+28.1\%}}$ & \textbf{60.36}$_{\textcolor{mygreen}{+7.2\%}}$ & 67.34$_{\textcolor{red}{-0.1\%}}$ & 72.4$_{\textcolor{red}{-0.01\%}}$ & \textbf{89.7}$_{\textcolor{mygreen}{+0.003\%}}$ \\
\ours$_{(Full)}$ & 20.21$_{\textcolor{mygreen}{+24.4\%}}$ & 60.24$_{\textcolor{mygreen}{+7.0\%}}$ & 66.21$_{\textcolor{red}{-1.8\%}}$  & 72.0$_{\textcolor{red}{-0.01\%}}$ & 89.5$_{\textcolor{mygreen}{+0.001\%}}$ \\
\bottomrule
\end{tabular}
}
\caption{\textbf{Results on Egocentric and Exocentric benchmarks.} Finetuning on \traindata\ benefits other egocentric benchmarks such as EgoTempo and EgoIllusion, achieving up to 28.1\% accuracy gain. Our model also maintains strong performance on exocentric video QA benchmarks such as VideoMME and AVQA.}\label{tab:ego_exo_comparison}
\end{table}
To further evaluate the generalizability of our fine-tuned model, we report its performance on additional egocentric video–language benchmarks, including EgoTempo, EgoSchema, and EgoIllusion. As shown in Table~\ref{tab:ego_exo_comparison}, fine-tuning on \traindata\ leads to notable accuracy improvements on EgoTempo and EgoIllusion, with gains of up to 28.1\%, while maintaining competitive performance on EgoSchema, showing only a marginal decrease of 0.1\%. Note that LoRA performs slightly better than full finetuning in terms of improving other datasets. These results demonstrate that training on \traindata\ enhances audio-visual understanding without causing overfitting, and further complements performance across diverse egocentric video QA tasks. We also evaluate our finetuned model on non-egocentric datasets.

In addition to egocentric benchmarks, we evaluate our fine-tuned model on popular exocentric Video QA benchmarks, including VideoMME (Short duration split w/o subtiles)~\citep{fu2024video} and AVQA~\citep{yang2022avqa}. As shown in Table~\ref{tab:ego_exo_comparison}, despite being fine-tuned exclusively on egocentric QAs, our model almost retains its original performance on VideoMME and slightly outperforms the base model Qwen2.5-Omni on audio-visual QAs in AVQA.






\begin{figure*}[t]
\centering
\begin{minipage}[t]{0.48\textwidth}
\centering
\includegraphics[width=\columnwidth]{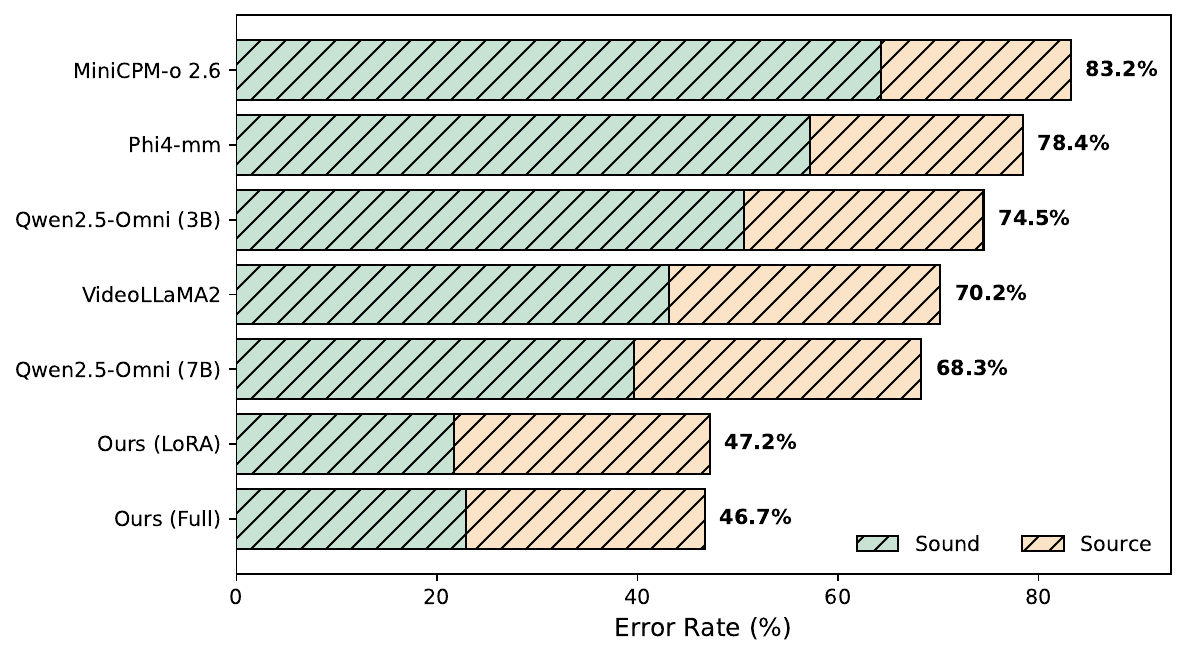}
\caption{\textbf{Error Analysis on Sound-Source Association (SSA).}}
\label{fig:SSA_error}
\end{minipage}
\hfill
\begin{minipage}[t]{0.48\textwidth}
\centering
\includegraphics[width=\columnwidth]{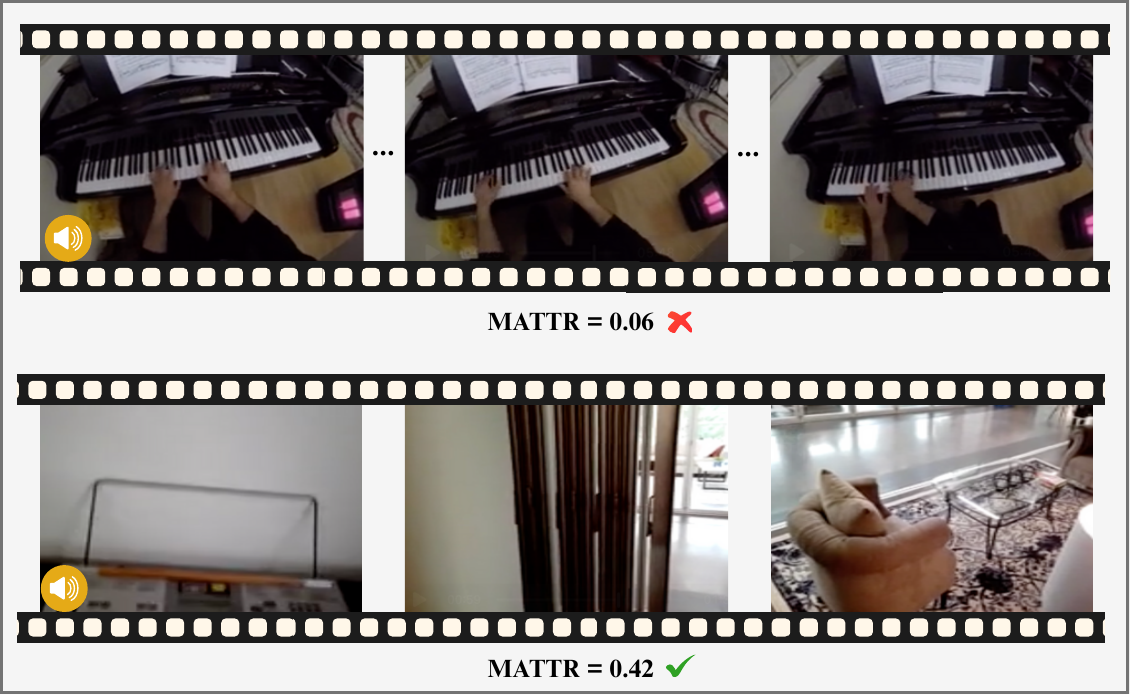}
\caption{\textbf{Examples of video filtering based on MATTR scores.}}
\label{fig:mattr_example}
\end{minipage}
\end{figure*}

\subsection{Error Analysis}
Intrigued by the weak performance of current MLLMs across all tasks in \evaldata, we conduct a detailed error analysis separately for the close-ended and open-ended tasks to analyze their behavior patterns. Our key findings are summarized below:

\noindent{\textbf{Close-Ended Tasks.}} Firstly, as illustrated in Table~\ref{tab:avh_comparison} and~\ref{tab:av_mcq_comparison}, for close-ended tasks such as Audio-Visual Hallucination (AVH) and Temporal Reasoning (TR), we evaluate MLLMs ability to independently perceive multisensory inputs within egocentric videos, specifically, human actions, visual objects, and sounds, including both foreground and background audio cues. This is achieved by separately evaluate the accuracy on the 3 subsets of corresponding tasks. Overall, our analysis reveals a consistent trend: \textit{MLLMs struggle the most with identifying sounds, followed by human actions, while performing relatively better at recognizing visual objects.} For instance, in the TR task, the best-performing model, Qwen2.5-Omni, achieves only 36.1\% accuracy in identifying sounds, exhibiting substantial performance gaps of 28.5\% and 7.4\% relative to visual object and human action identification, respectively. Furthermore, we find that model fine-tuned on our dataset, achieves a significant boost in its ability to independently perceive multisensory inputs within egocentric videos. For example, in the AVH task, compared to Qwen2.5-Omni, our fine-tuned model shows a reduced hallucination rate of 15.9\%, 11.0\%, and 30.0\% when identifying human actions, visual objects, and sounds, respectively.

\noindent{\textbf{Open-Ended Tasks.}} We further extend our analysis to open-ended tasks such as Source-Sound Association (SSA) in EgoAVUBench, aiming to determine which modality contributes more to the overall error rate when models are asked to produce joint audio–visual descriptions. Specifically, we randomly sample 200 data points from the SSA task and, for each incorrect response across different MLLMs, manually annotate whether the error stems from inaccurate sound perception or an incorrect source description (which may involve visible objects or human–object interactions in the egocentric video). As shown in Fig.~\ref{fig:SSA_error}, our model achieves a substantially lower error rate of 21.1\% compared to the next best model, Qwen2.5-Omni. Interestingly, other MLLMs such as MiniCPM-o and Phi4-mm, which exhibit much higher error rates, show that over 72\% of their errors stems from incorrect or missed sound descriptions rather than misidentified human–object interactions. This indicates that \textit{MLLMs primarily fail due to their limited ability to accurately perceive and interpret sounds, leading to incorrect associations with their visible sources.}

\section{Conclusion}\label{sec:conclusion}


We presented \methodname, a novel data engine to enhance egocentric audio-visual understanding by addressing the data limitations. \methodname~employed modularized MLLMs to enhance egocentric narrations, and leveraged multi-modal context graphs to generate diverse, high-quality audio-visual QA pairs. Our evaluation benchmark \evaldata\ revealed for the first time that existing MLLMs exhibit consistent vision bias, often neglecting or hallucinating audio information in egocentric videos. Our large-scale training dataset \traindata\ effectively mitigated this gap, significantly improved performance on both \evaldata\ and existing egocentric benchmarks. Importantly, \methodname\ demonstrated the self-learning potential of MLLMs: using uni-modal capabilities to improve the joint-modal capability. In terms of limitation, our training data, though with carefully designed filtering techniques, still contains noise from open source MLLM outputs. We believe this problem can be continually alleviated as the uni-modal capabilities of MLLMs improve, and leave the development of more soficsticated approaches as future work.

\clearpage
\newpage
\bibliographystyle{assets/plainnat}
\bibliography{paper}

\clearpage
\newpage
\beginappendix

\section{Additional Details on \methodname}\label{appdx:video_filter}

\subsection{Prompts}
\noindent{\textbf{Prompt Used.}} We describe the various prompts used in \methodname. These prompts cover the generation of multiple modules including the Multi-modal Context Graph (see Fig.~\ref{fig:supp_prompt}), Audio-Visual Narration (see Fig.\ref{fig:supp_prompt_narration}), and task-specific QA generation (see Fig.~\ref{fig:supp_sound_prompt}, Fig.~\ref{fig:halu_sound}, Fig.~\ref{fig:halu_action}, Fig.~\ref{fig:halu_object},
Fig.~\ref{fig:supp_temporal_prompt},
Fig.~\ref{fig:supp_temporal_mcq_prompt}, Fig.~\ref{fig:supp_dense_narration_prompt}). The general structure of each prompt includes: (i) defining the objective, (ii) specifying the input format, (iii) describing the task, (4) providing general instructions, (iv) including human-generated examples to enable in-context learning, and (v) specifying the output format.

\subsection{MATTR}
For video filtering, we utilize the Moving-Average Type-Token Ratio (MATTR) to identify videos with rich multimodal diversity. We set the window size to $w=200$ tokens based on the average token length of combined narrations across video segments in our dataset, ensuring the window captures sufficient multi-modal context for meaningful diversity measurement. We retain videos whose MATTR exceeds $\tau = 0.3$, effectively removing the bottom 25\% of our distribution to filter out static or repetitive descriptions. This threshold was determined through manual inspection of 100 randomly sampled videos across different MATTR ranges. Videos below $\tau = 0.3$ predominantly featured repetitive actions with limited object diversity and minimal auditory variation, while videos above this threshold exhibited richer multimodal dynamics (refer to Fig~\ref{fig:mattr_example} for more examples).

\subsection{Ablation on Multi-Modal Context Graphs}\label{appdx:MCG_ablation}
\begin{figure*}
    \centering
    \includegraphics[width=1.0\textwidth]{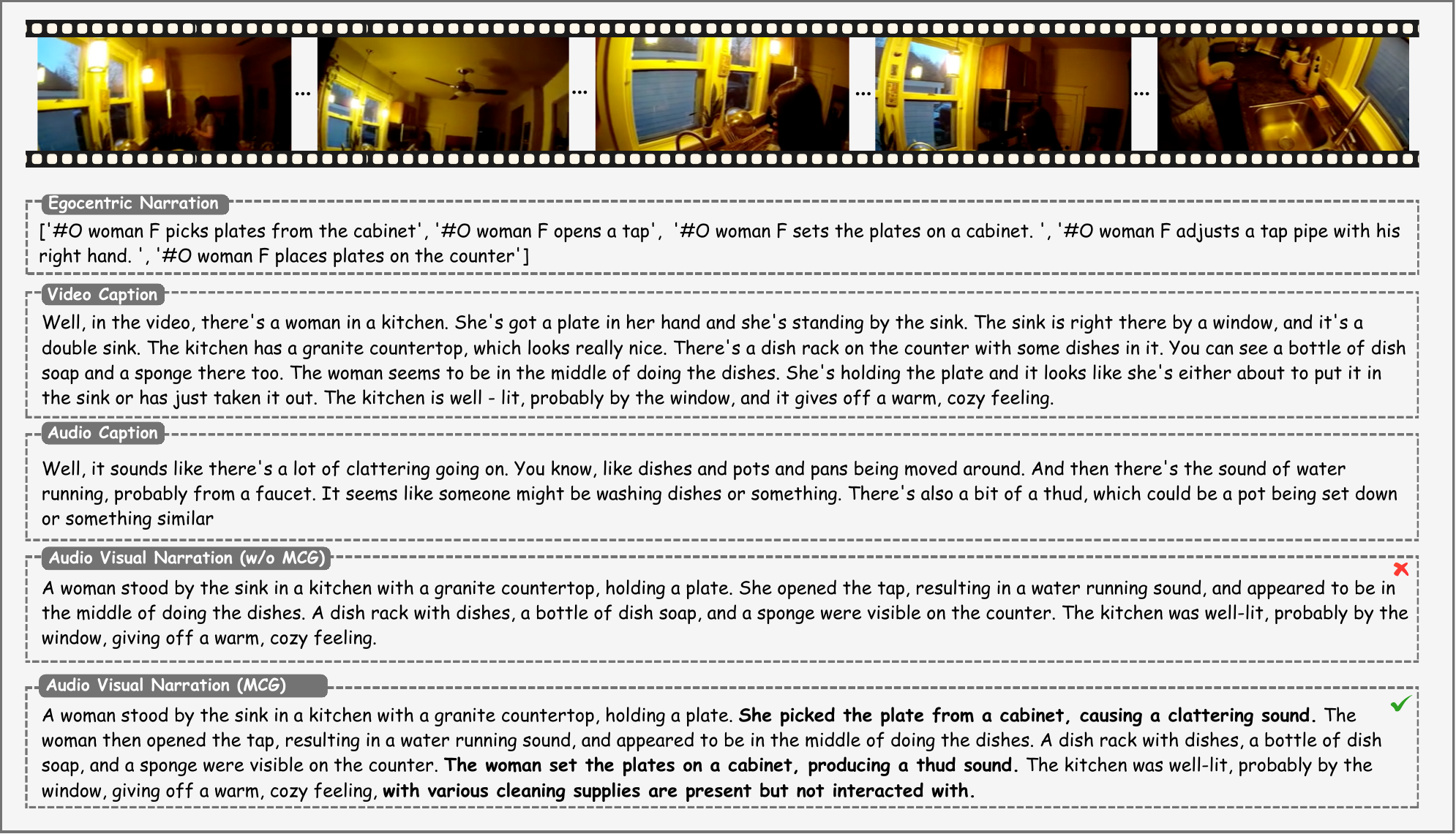}
    \caption{\textbf{Qualitative comparison of audio-visual narrations generated with and without (w/o) Multi-modal Context Graph (MCG).} Our MCG-based approach produces narrations with superior audio-visual coherence, accurately capturing action sequences and sound-source associations, while the direct method (w/o MCG) often misses critical sounds or action sequences.}
    \label{fig:mcg_qualitative}
\end{figure*}

To validate the necessity of the Multi-Modal Context Graphs (MCG) component in \methodname, we conducted an ablation experiment on 200 randomly sampled video clips. We compared our MCG-based pipeline against a direct baseline where LLaMA-3-70B generates audio-visual narrations directly from enhanced narrations (video caption, image caption, audio caption, and action narration) without the intermediate MCG structure. We manually evaluated both output, assessing: (1) completeness of sound-source associations, (2) accuracy of action sequences, and (3) overall audio-visual coherence. We observed that the direct method produced errors in 82 out of 200 captions (41.0\%), with the breakdown as follows: 48 captions (19.0\%) missed or incorrectly associated sound sources, 31 captions (15.5\%) omitted crucial action sequences or interaction details, and 17 captions (3.5\%) exhibited both issues (refer to Fig.~\ref{fig:mcg_qualitative} for example). In contrast, the our MCG-based approach reduced errors to 21 out of 200 captions (10.5\%), representing a 76.1\% relative error reduction.

\section{Manual Effort for \evaldata\ Construction}
To ensure the reliability of \evaldata, we conducted extensive manual verification across all 3,000 question-answer pairs covering 900 egocentric videos. Each video was carefully reviewed by trained annotators, taking approximately 2-3 minutes per video to verify temporal alignment and audio-visual correspondence. Out of 3,000 QA pairs, 1,524 pairs (50.8\%) were modified or corrected during this process. For open-ended tasks (SSA, AVDN, AVSN), corrections primarily addressed missing sounds, incorrect human-object interactions, and sound-source misalignments to ensure accurate audio-visual grounding. For close-ended tasks (TR, AVH), we verified answer correctness and enhanced distractor quality by ensuring multiple-choice options were sufficiently challenging and plausible while avoiding options that were too similar to correct answers or obviously incorrect. The complete manual verification process required approximately 225 hours of human annotation effort.

\section{Additional Details on Narration Enhancement}
\label{appdx:MLLM_hallucination}
\noindent{\textbf{Prompts.}} To capture spatial details, we extract the center frame and prompt Qwen2.5-VL with \textit{\textbf{``Identify all the objects visible in the image''}} to detail all objects present in the video clip. To capture temporal dynamics, we utilize Qwen2.5-Omni in two stages: first, we process only the video frames with the prompt \textbf{\textit{``Describe the video in detail''}} to capture visible activities; then, we process the audio with the prompt \textit{\textbf{``Describe all the sounds heard in detail''}} to capture auditory information.


\section{Additional Experiment Details}\label{appdx:implementation}
\noindent{\textbf{Training Details.}}
For both LoRA and full fine-tuning, we use a maximum context length of 30{,}000 tokens and sample videos at 1~FPS with a frame resolution of $256 \times 256$. Training is performed using DeepSpeed ZeRO-3 with a learning rate of $1 \times 10^{-4}$ and a cosine schedule with 10\% warmup over 5 epochs. We perform balanced sampling, i.e, sample with equal weights from each task, during training.

\noindent{\textbf{Evaluation Details.}} Fig.~\ref{fig:supp_grader_prompt} presents the prompt used for our \textit{LLM-as-judge} evaluation. Following prior work~\citep{plizzari2025omnia}, we assess the reliability of LLM-based scoring by measuring its alignment with human judgments on 300 randomly sampled open-ended QA pairs from \evaldata. The resulting human-alignment rate is 87.6\%, indicating strong alignment between the two.



\begin{figure*}[t]
\centering

\begin{tcolorbox}[
    width=\textwidth,
    colback=gray!10,
    colframe=black!20,
    boxrule=0.3pt,
    arc=2pt,
    left=6pt,
    right=6pt,
    top=6pt,
    bottom=6pt
]
\small
\ttfamily
\textbf{Objective:} You are an AI assistant tasked with performing a high-fidelity analysis of video content. Your role is to function as an evidence extractor, not an open-world reasoner. You must strictly use the provided captions to identify object interactions and to analyze sounds, grounding every piece of information directly to the source text.

\textbf{Inputs:} You will be provided with a JSON object containing the following four keys:
\begin{itemize}
    \item \textit{Video Caption:} Describes the overall visual scene, including events and actions of entities other than the narrator (e.g., animals, other people, environmental events).
     \item \textit{Image Caption:} Describes the diverse objects visible in the center frame.
    \item \textit{Audio Caption:}  A transcript of sounds and audio events.
    \item \textit{Action Narration:} Describes the specific actions performed by the primary person ($\# C C$ denotes person)
    
\end{itemize}

\textbf{Task:} Analyze the provided captions to generate a structured multimodal context graph in the form of JSON that captures multimodal relationship. For generating the multimodal context graph, follow the instruction mentioned below:

\textbf{Instructions:}
\begin{itemize}
    \item \textit{Identify Interacted Objects:} Parse the action narration to find all objects the primary person is described as touching, holding, using, or manipulating. Compile these into the "interacted objects" list with what action the person performed.
    \item \textit{Identify Background Objects:} Take the complete list of objects from image caption. Create a new list containing only the items from image caption that are NOT in your "interacted objects" list. This will be your "non interacted objects" list.
    \item \textit{Identify Sound-Source Associations} This is a strict, evidence-based process. Ideally there can be two type of sound, sound caused by action/object or background sound. Your task is to capture both of them for audio caption

\noindent A. Find the Grounding Evidence for foreground sound: For each sound in audio caption, you must search action narration and video caption to find the specific text that describes the action or event causing it. Look in action narration for causes related to the primary person's actions.
Look in video caption for causes related to other entities (animals, other people) or general scene events. ambient sound include music, background noise, etc.
Crucially: If no direct textual evidence can be found in either caption that explains the sound's origin, the sound is a background sound.

\noindent B. Exclude "Unquestionable" Sounds: Even if grounded, do NOT include a sound if it falls into these categories:
Mundane Biological Sounds: Common sounds like "breathing," "sighing," "swallowing."
Vague Ambient Noise: like "white noise" or "faint hum."

\noindent C. Determine the sound category: Classify as Foreground Sound (from the human action or visible object) or Background Sound.

\noindent D. Handle Empty Results: If no sounds pass the filtering and grounding process, the "sounds" list in your output must be an empty list ([]).
\end{itemize}

\textbf{Human Generated Examples}: Here are 5 human created examples for the correct execution <examples>.

\textbf{Important Note:}
\begin{itemize}
    \item In the example how "giggle" was excluded because it had no grounding evidence in video caption or action narration.
    \item Source description will contain either the corresponding action or the object that can produced the sound
    \item Sound category can have foreground sound such as ``Action Sound'', ``Object Sound'' or background such as ``Ambient Sound''
    
\end{itemize}
\textbf{Final Output Format:} Your entire response MUST be a single, valid JSON object following the structure of the example. Do not include any text outside of the JSON structure.

Here is the input: <input>
\end{tcolorbox}

\caption{\centering\textbf{Prompt For Generating Multi-Modal Context Graphs.}}
\label{fig:supp_prompt}
\end{figure*}
\begin{figure*}[t]
\centering
\begin{tcolorbox}[
    width=\textwidth,
    colback=gray!10,
    colframe=black!20,
    boxrule=0.3pt,
    arc=2pt,
    left=6pt,
    right=6pt,
    top=6pt,
    bottom=6pt,
]
\small
\begin{verbatim}
{
        "interacted_objects": [
            ["sink", "#C C rinses both hands"],
            ["tap", "#C C turns on tap"],
            ["door", "#C C opens the door"],
        ],
        "background_objects": [
            "oranges",
            "sponge",
            "red chair",
            "microwave",
            "cabinets"
        ],
        "sounds": [
            {
                "acoustic_description": "water flowing sound",
                "source": "#C C turns on tap",
                "evidence_source": "action_narration",
                "sound_category": "Foreground Sound"
            },
            {
                "acoustic_description": "hands being rinsed sound",
                "source": "#C C rinses both hands",
                "evidence_source": "action_narration",
                "sound_category": "Foreground Sound"
            },
            {
                "acoustic_description": "door opening and closing sound",
                "source": "#C C opens the door",
                "evidence_source": "action_narration",
                "sound_category": "Foreground Sound"
            }
        ]
    }
\end{verbatim}
\end{tcolorbox}
\caption{\centering\textbf{Example of MCG.} Example of MCG generated in JSON format using the above-mentioned prompt.}
\label{fig:raw_json_example}
\end{figure*}
\begin{figure*}[t]
\centering

\begin{tcolorbox}[
    width=\textwidth,
    colback=gray!10,
    colframe=black!20,
    boxrule=0.3pt,
    arc=2pt,
    left=6pt,
    right=6pt,
    top=6pt,
    bottom=6pt
]
\small
\ttfamily

\textbf{Objective:} Your job is to generate a single, detailed, and objective paragraph summarizing what can be seen and heard in the video clip. 

\textbf{Input:} You will be given:

\begin{itemize}
    \item A free-form natural language paragraph summarizing what happens in the video. This is typically derived from a loose transcription or human description of the scene.
    \item A list of short, possibly overlapping or action tags extracted from the video. These may contain minor inconsistencies, but offer clues about human interactions and movements in the scene. Focus closely on the interaction starting with \#C C.
    \item A multi-modal context graph represented as a structured JSON object containing: 
\texttt{"interacted\_objects"} --- objects the person interacted with, 
\texttt{"non\_interacted\_objects"} --- objects present but not interacted with, and 
\texttt{"sounds"} --- sound events grounded in the narration, with descriptions and causes.
\end{itemize}

\textbf{Task:} Write a single, coherent paragraph that summarizes the video scene in detail, following these rules:

\textbf{Instructions}

\begin{itemize}
    \item Clearly describe all key actions in the scene, combining the raw description and narration tags.
    \item Include all interacted objects, and describe how each was interacted with.
    \item Mention all non-interacted objects that are visible or relevant once, integrated naturally into the scene description. Do not repeat them again at the end.
    \item Integrate sound events by describing what caused them and when, grounded in the referenced actions.
    \begin{itemize}
        \item Not all actions have corresponding sounds.
        \item Only include sounds that are listed in the scene graph.
        \item Use semantically appropriate or naturalistic descriptions for acoustic events. For instance, if the sound is caused by shaking a spray bottle, you may refer to it as ``crunching or rattling''.
    \end{itemize}
    \item Use an objective and factual tone—avoid any emotional, subjective, or evaluative language (e.g., no ``cute,'' ``interesting,'' or ``simple'').
    \item Write in past tense.
    \item Ensure the paragraph flows naturally and avoids redundancy.
\end{itemize}

\textbf{Human Generated Examples}: Here are 5 human created examples for the correct execution <examples>.

\textbf{Final Output Format:} The final output must be in JSON format with key as "caption"\\

Here is the input to generate the caption: <input>

\end{tcolorbox}

\caption{\centering\textbf{Prompt For Generating Audio-Visual Narration.}}
\label{fig:supp_prompt_narration}
\end{figure*}

\begin{figure*}[t]
\centering
\begin{tcolorbox}[
    width=\textwidth,
    colback=gray!10,
    colframe=black!20,
    boxrule=0.3pt,
    arc=2pt,
    left=6pt,
    right=6pt,
    top=6pt,
    bottom=6pt
]
\small
\ttfamily

\textbf{Objective:} You are an AI assistant tasked with analyzing a video segment and performing three tasks:
\begin{itemize}
    \item Generate a single open-ended question about the sound-source association observed in the video.
    \item Produce a natural, human-like narration that links sounds to the actions and objects responsible.
    \item Generate a detailed, structured answer to the question, grounded entirely in the provided scene graph metadata.
\end{itemize}

\textbf{Input:} You will be provided with:
\begin{itemize}
    \item video description: description of the video segment
    \item Multi-modal Context Graph: <Details on Multi-modal Context Graph>
\end{itemize}

\textbf{Instructions:} Follow the below instruction to complete the task:
\begin{itemize}

    \item \textit{Question Generation.} If the "sounds" list is empty or missing, return this exact string as the only output: "No significant sound is present in the video clip." Otherwise, use the template below: <template>

    \item When narrating egocentric data, a person is sometimes referred to by capital letters, such as "C." When writing the description, treat such IDs as referring to a person. For example, if a sound-producing evidence states "person C is clapping," it should be treated as "the person is clapping."

    \item \textit{Detailed Answer Generation.}
    \begin{itemize}
        \item Structure the answer as follows: Begin with a sentence that clearly states how many distinct grounded sound events were present. Then provide one sentence for each sound, explaining what caused it by using the acoustic description and grounding evidence. Treat C as the person in the video.
        \item Do not speculate or add interpretation beyond the metadata.
        \item Do not include any text outside of the JSON structure.
        \item Do not include any step by step explanations.
    \end{itemize}

\end{itemize}

\textbf{Human Generated Examples}: Here are 5 human created examples for the correct execution <examples>.

\textbf{Output Format:} Output must be in a JSON format with following key "question" and "answer".

Here is the input to generate the question-answer pair: <input>

\end{tcolorbox}

\caption{\centering\textbf{Prompt for generating Sound-Source Association Question-Answer pair.}}
\label{fig:supp_sound_prompt}
\end{figure*}

\begin{figure*}[t]
\centering
\begin{tcolorbox}[
    width=\textwidth,
    colback=gray!10,
    colframe=black!20,
    boxrule=0.3pt,
    arc=2pt,
    left=6pt,
    right=6pt,
    top=6pt,
    bottom=6pt
]
\small
\ttfamily

\textbf{Objective:} Generate two sound-related question--answer pairs from an egocentric video caption that describes a person's visible actions, sounds, and objects. The output should be formatted as JSON with one correct and one hallucinated sound question.

\textbf{Input:} You will be given an egocentric video narration containing descriptions of:
\begin{itemize}[noitemsep]
    \item The person's visible actions
    \item Distinctive sounds (e.g., hissing, tapping, scraping)
    \item Objects present in the scene
    \item Temporal information about when events occur
\end{itemize}

\textbf{Instructions:} Follow the instruction below:
\begin{itemize}
    \item \textit{Focus on distinctive sounds} such as foreground sounds related to human-object interaction such as hissing, tapping etc. or background sounds such as bird chirping etc.
    \item \textit{Generate one correct question:} Ask about a sound explicitly mentioned in the narration.
    \item \textit{Generate one hallucinated question}: Ask about a plausible sound that is \textit{not} mentioned in the narration.
    \item \textit{Answer format}: Answers must be in binary format "Yes" or "No".
\end{itemize}

\textbf{Output Format:} The output must be in JSON format with following keys: "question", "question type" including "Factual", "Hallucinated" and "answers".

Here is the input to generate the question-answer pair: <input>
\end{tcolorbox}

\caption{\centering\textbf{Prompt for generating Audio-Visual Hallucination (Sound) Question-Answer pair.}}
\label{fig:halu_sound}
\end{figure*}

\begin{figure*}[t]
\centering
\begin{tcolorbox}[
    width=\textwidth,
    colback=gray!10,
    colframe=black!20,
    boxrule=0.3pt,
    arc=2pt,
    left=6pt,
    right=6pt,
    top=6pt,
    bottom=6pt
]
\small
\ttfamily

\textbf{Objective:} Generate two action-related question--answer pairs from an egocentric video caption that describes a person's visible actions, sounds, and objects. The output should be formatted as JSON with one correct and one hallucinated action question.

\textbf{Input:} You will be given an egocentric video narration containing descriptions of:
\begin{itemize}[noitemsep]
    \item The person's visible actions
    \item Distinctive sounds (e.g., hissing, tapping, scraping)
    \item Objects present in the scene
    \item Temporal information about when events occur
\end{itemize}

\textbf{Instructions:} Follow the instruction below:
\begin{itemize}
    \item \textit{Focus on distinctive, non-trivial actions} such as wiping, twisting, or squeezing. Avoid trivial actions such as breathing, walking, or placing.
    \item \textit{Generate one correct question:} Ask about an action explicitly mentioned in the narration.
    \item \textit{Generate one hallucinated question}: Ask about a plausible action that is \textit{not} mentioned in the narration.
    \item \textit{Answer format}: Answers must be in binary format "Yes" or "No".
\end{itemize}

\textbf{Output Format:} The output must be in JSON format with following keys: "question", "question type" including "Factual", "Hallucinated" and "answers".

Here is the input to generate the question-answer pair: <input>
\end{tcolorbox}

\caption{\centering\textbf{Prompt for generating  Audio-Visual Hallucination (Action) Question-Answer pairs.}}
\label{fig:halu_action}
\end{figure*}

\begin{figure*}[t]
\centering
\begin{tcolorbox}[
    width=\textwidth,
    colback=gray!10,
    colframe=black!20,
    boxrule=0.3pt,
    arc=2pt,
    left=6pt,
    right=6pt,
    top=6pt,
    bottom=6pt
]
\small
\ttfamily

\textbf{Objective:} Generate two object-related question--answer pairs from an egocentric video caption that describes a person's visible actions, sounds, and objects. The output should be formatted as JSON with one correct and one hallucinated object question.

\textbf{Input:} You will be given an egocentric video narration containing descriptions of:
\begin{itemize}[noitemsep]
    \item The person's visible actions
    \item Distinctive sounds (e.g., hissing, tapping, scraping)
    \item Objects present in the scene
    \item Temporal information about when events occur
\end{itemize}

\textbf{Instructions:} Follow the instruction below:
\begin{itemize}
    \item \textit{Focus on specific, manipulable objects.} Avoid generic nouns like 'things', 'stuff', or 'material'.
    \item \textit{Generate one correct question:} Ask about an object explicitly mentioned in the narration.
    \item \textit{Generate one hallucinated question}: Ask about a plausible object that is \textit{not} mentioned in the narration.
    \item \textit{Answer format}: Answers must be in binary format "Yes" or "No".
\end{itemize}

\textbf{Output Format:} The output must be in JSON format with following keys: "question", "question type" including "Factual", "Hallucinated" and "answers".

Here is the input to generate the question-answer pair: <input>
\end{tcolorbox}

\caption{\centering\textbf{Prompt for generating Audio-Visual Hallucination (Object) Question-Answer pairs.}}
\label{fig:halu_object}
\end{figure*}

\begin{figure*}[t]
\centering
\begin{tcolorbox}[
    width=\textwidth,
    colback=gray!10,
    colframe=black!20,
    boxrule=0.3pt,
    arc=2pt,
    left=6pt,
    right=6pt,
    top=6pt,
    bottom=6pt
]
\small
\ttfamily

\textbf{Objective:} Generate two temporal reasoning question--answer pairs from a list of chronological video narrations, focusing on the order of Action, Object, and Sound events. The output should be formatted as a JSON list containing one "before" and one "after" question.

\textbf{Input:}
\begin{itemize}
    \item A list of narration describing what happens in the video in chronological order (\texttt{\{caption\_list\}}).
    \item The specific question type to be generated (\texttt{\{type\}}: one of Action-Action, Action-Object, or Action-Sound).
\end{itemize}

\textbf{Instructions:} Follow the steps below:
\begin{enumerate}
    \item \textit{Identify Distinct Events:} Identify several unique, non-trivial, and non-repetitive events, each describing an \textbf{Action}, an \textbf{Object}, or a \textbf{Sound}.
    \item \textit{Select Event Pair (E1, E2):} Choose two events occurring at different times that match the required category (\texttt{\{type\}}). E1 must chronologically precede E2.
    \item \textit{Generate Questions:} Create one "before" question (referencing E2) and one "after" question (referencing E1) using the corresponding template:
        \begin{itemize}
            \item \textbf{Action--Action}
            \begin{itemize}
                \item Before: "What action was the person performing before \texttt{<E2>}?"
                \item After: "What action did the person perform after \texttt{<E1>}?"
            \end{itemize}
            \item \textbf{Action--Object}
            \begin{itemize}
                \item Before: "What objects can be seen before the person performed the \texttt{<E2>} action?"
                \item After: "What objects can be seen after the person performed the \texttt{<E1>} action?"
            \end{itemize}
            \item \textbf{Action--Sound}
            \begin{itemize}
                \item Before: "What sound can be heard before the person \texttt{<E2>}?"
                \item After: "What sound can be heard after the person \texttt{<E1>}?"
            \end{itemize}
        \end{itemize}
    \item \textit{Answer and options:} Write a concise, naturalistic answer as if you watched the video. Include three plausible \texttt{options} that fit the context but are temporally incorrect.
\end{enumerate}

\textbf{Output Format:} The output must be a JSON list of exactly two question objects (one "before" and one "after") with the following keys: "question", "answer", "type", and "options".
\end{tcolorbox}

\caption{\textbf{Prompt for generating Temporal Reasoning (before/after) Question-Answer Pairs}}
\label{fig:supp_temporal_prompt}
\end{figure*}

\begin{figure*}[t]
\centering
\begin{tcolorbox}[
    width=\textwidth,
    colback=gray!10,
    colframe=black!20,
    boxrule=0.3pt,
    arc=2pt,
    left=6pt,
    right=6pt,
    top=6pt,
    bottom=6pt
]
\small
\ttfamily

\textbf{Objective:} Generate one multiple-choice question about the temporal order of four events derived from a sequence of chronological egocentric video narration.

\textbf{Input:} A sequence of detailed narration in chronological order describing what happens in a egocentric video.

\textbf{Instructions:} Follow the steps below:
\begin{itemize}
    \item \textit{Identify Four Grounded Events:} Identify four unique, non-trivial events. Each event must include concise but meaningful details about the person's activity, the visible surroundings, and any sounds mentioned.
    \item \textit{Grounding Constraint:} All events must be directly derived from the narrations. Do not hallucinate or invent any objects, sounds, or actions.
    \item \textit{Create Temporal Question:} Create one general multiple-choice question that asks about the temporal order of the four events (e.g., "Which event happened first?", "Which moment occurred last?").
    \item \textit{Create Options:} List the four events as options A, B, C, and D. Ensure the description for each option is the exact description provided in the \texttt{events} list.
    \item \textit{Provide Correct Answer:} Indicate the correct temporal order by selecting one of the options (A, B, C, or D) as the \texttt{correct\_answer}.
\end{itemize}

\textbf{Output Format:} The output must be in JSON format with the following keys: "events" (a list of the four descriptions), "question", "options" (a map of A, B, C, D to the event descriptions), and "answer".
\end{tcolorbox}

\caption{\textbf{Prompt for generating Temporal Reasoning (Event Ordering) Question-Answer Pairs}}
\label{fig:supp_temporal_mcq_prompt}
\end{figure*}
\begin{figure*}[t]
\centering
\begin{tcolorbox}[
    width=\textwidth,
    colback=gray!10,
    colframe=black!20,
    boxrule=0.3pt,
    arc=2pt,
    left=6pt,
    right=6pt,
    top=6pt,
    bottom=6pt
]
\small
\ttfamily

\textbf{Objective:} Write a single, coherent, dense narration summarizing the entire video based on a list of 10-second captions.

\textbf{Input:} A list of narration, each describing a 10-second segment of a egocentric video, including \texttt{start\_time}, \texttt{end\_time}, and the \texttt{caption} text.

\textbf{Instructions:} The final output must be a single, fluent paragraph that acts as a dense narration. The paragraph must adhere to the following rules:
\begin{itemize}
    \item Integrate all actions, objects, and sounds across the full video.
    \item Use timestamps in seconds to indicate when key events occurred.
    \item Group similar or adjacent events into continuous spans.
    \item Avoid listing or repeating captions verbatim.
    \item Use only the information in the input captions.
    \item Be concise and fluent.
    \item Do not invent any new information or context.
\end{itemize}

\textbf{Human Generated Examples:} Here are 3 human created dense narration: <examples> 

\textbf{Output Format:} A single paragraph, not a JSON object.

Here is the input: <input>
\end{tcolorbox}
\caption{\textbf{Prompt for generating Audio Visual Dense Narration}}
\label{fig:supp_dense_narration_prompt}
\end{figure*}
\begin{figure*}[t]
\centering
\begin{tcolorbox}[
    width=\textwidth,
    colback=gray!10,
    colframe=black!20,
    boxrule=0.3pt,
    arc=2pt,
    left=6pt,
    right=6pt,
    top=6pt,
    bottom=6pt
]
\small
\ttfamily

\textbf{Objective:} Act as an impartial grader to evaluate a \texttt{PREDICTED\_ANSWER} against a \texttt{GROUNDING\_ANSWER} with respect to a \texttt{QUESTION}.

\textbf{Input:}
\begin{itemize}[noitemsep]
    \item \texttt{QUESTION}: The question posed to the model.
    \item \texttt{GROUNDING\_ANSWER}: The authoritative reference answer.
    \item \texttt{PREDICTED\_ANSWER}: The model's answer to be graded.
\end{itemize}

\textbf{Instructions for Grading:}
\begin{enumerate}
    \item \textit{Comparison:} Compare \texttt{PREDICTED\_ANSWER} to \texttt{GROUNDING\_ANSWER} with respect to the \texttt{QUESTION}.
    \item \textit{Assign Rating (1-5, integer only):}
        \begin{itemize}
            \item \textbf{5:} Fully correct, complete, and faithful to the grounding; no meaningful errors or omissions.
            \item \textbf{4:} Mostly correct; minor omissions or small inaccuracies that do not change the overall correctness.
            \item \textbf{3:} Partially correct; captures some key points but misses important details or includes notable inaccuracies.
            \item \textbf{2:} Largely incorrect; substantial errors, contradictions, or missing major required points.
            \item \textbf{1:} Incorrect/irrelevant; contradicts the grounding or fails to answer the question.
        \end{itemize}
    \item \textit{Provide Reasoning:} Briefly explain the rating (1–4 concise sentences).
\end{enumerate}

\textbf{Judging Rules (Priorities):}
\begin{itemize}
    \item Prioritize factual alignment with the \texttt{GROUNDING\_ANSWER}. Contradictions result in heavy penalization.
    \item Extra details are acceptable only if they do not conflict with the grounding and remain relevant to the \texttt{QUESTION}.
    \item Penalize hallucinations, unverifiable claims, safety issues, and failure to address the core of the \texttt{QUESTION}.
    \item Do not reward verbosity or style unless it improves factual accuracy or completeness with respect to the grounding.
    \item If the grounding indicates the question is unanswerable, judge whether the prediction correctly reflects that.
\end{itemize}

\textbf{Output Format:} The output \textbf{MUST} be valid JSON (no markdown, no extra text) with the following keys:
\begin{verbatim}
{
  "rating": <int value between 1 to 5>,
  "reason": "<string of 1-2 lines explaining the rating>"
}
\end{verbatim}
\end{tcolorbox}

\caption{\textbf{Prompt for LLM-as-judge evaluation}}
\label{fig:supp_grader_prompt}
\end{figure*}

\end{document}